\documentclass{article}
\usepackage{spconf,amsmath,graphicx, booktabs, xcolor}
\usepackage{amsmath,amssymb}
\usepackage{upgreek}

\renewcommand{\vec}[1]{\mathbf{#1}}
\newcommand{\mr}[1]{\mathrm{#1}}

\newcommand{\yy}{\vec{y}}
\newcommand{\xx}{\vec{x}}

\newcommand{\sss}{\vec{s}}
\newcommand{\xnorm}{\widehat{\vec{x}}}

\newcommand{\data}{\mathcal{D}}

\newcommand{\real}{\mathbb{R}}
\newcommand{\loss}{\mathcal{L}}
\newcommand{\lossSeg}{\loss_{\mr{seg}}}
\newcommand{\lossDis}{\loss_{\mr{dis}}}
\newcommand{\img}{\Upomega}

\usepackage{todonotes}

\title{ADVERSARIAL NORMALIZATION FOR MULTI DOMAIN IMAGE SEGMENTATION}
\name{Pierre-Luc Delisle, Benoit Anctil-Robitaille, Christian Desrosiers, Herve Lombaert}
\address{ETS Montreal, Canada}
\begin{document}

\maketitle

\begin{abstract}
 Image normalization is a critical step in medical imaging. This step is often done on a per-dataset basis, preventing current segmentation algorithms from the full potential of exploiting jointly normalized information across multiple datasets. 
 To solve this problem, we propose an adversarial normalization approach for image segmentation which learns common normalizing functions across multiple datasets while retaining image realism. The adversarial training provides an optimal normalizer that improves both the segmentation accuracy and the discrimination of unrealistic normalizing functions. Our contribution therefore leverages common imaging information from multiple domains. The optimality of our common normalizer is evaluated by combining brain images from both infants and adults. Results on the challenging iSEG and MRBrainS datasets reveal the potential of our adversarial normalization approach for segmentation, with Dice improvements of up to 59.6\% over the baseline. 
\end{abstract}

\section{Introduction}
\label{sec:intro}
In medical imaging applications, datasets with annotated images are rare and often composed of few samples. This causes an accessibility problem for developing supervised learning algorithms such as those based on deep learning. Although these algorithms have helped in automating image segmentation, notably in the medical field \cite{Kamnitsas2017,Dolz2019}, they need a massive number of training samples to obtain accurate segmentation masks that generalize well across different sites. One possible approach to alleviate this problem would be to use data acquired from multiple sites to increase the generalization performance of the learning algorithm. However, medical images from different datasets or sites can be acquired using various protocols. This leads to a high variance in image intensities and resolution, increasing the sensitivity of segmentation algorithms to raw images and thus impairing their performance. 

\begin{figure}[h]
    \includegraphics[width=\linewidth]{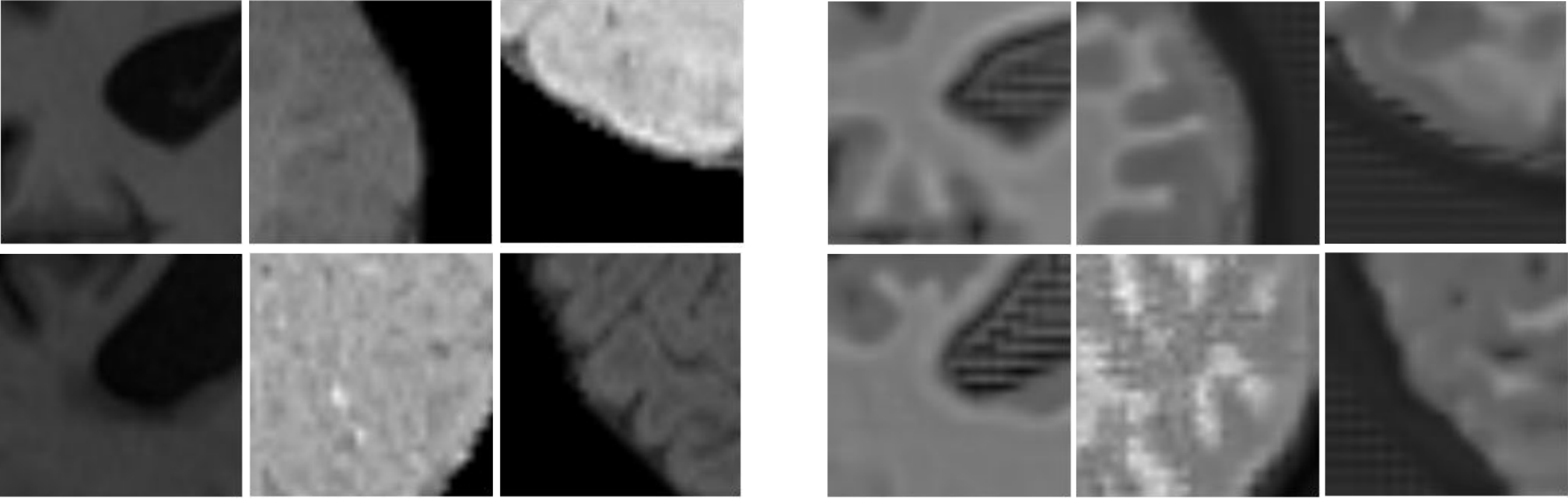}
    \caption{Mixed iSEG and MRBrainS inputs (\textbf{left}) and images generated with two pipelined FCNs without constraint on realism using only Dice loss (\textbf{right}). Images generated with only Dice loss preserve the structure required for segmentation but lack realism.}
    \label{fig1}
\end{figure}

\begin{figure*}[h]
\begin{center}
    \includegraphics[width=\linewidth, height=5.50cm, keepaspectratio]{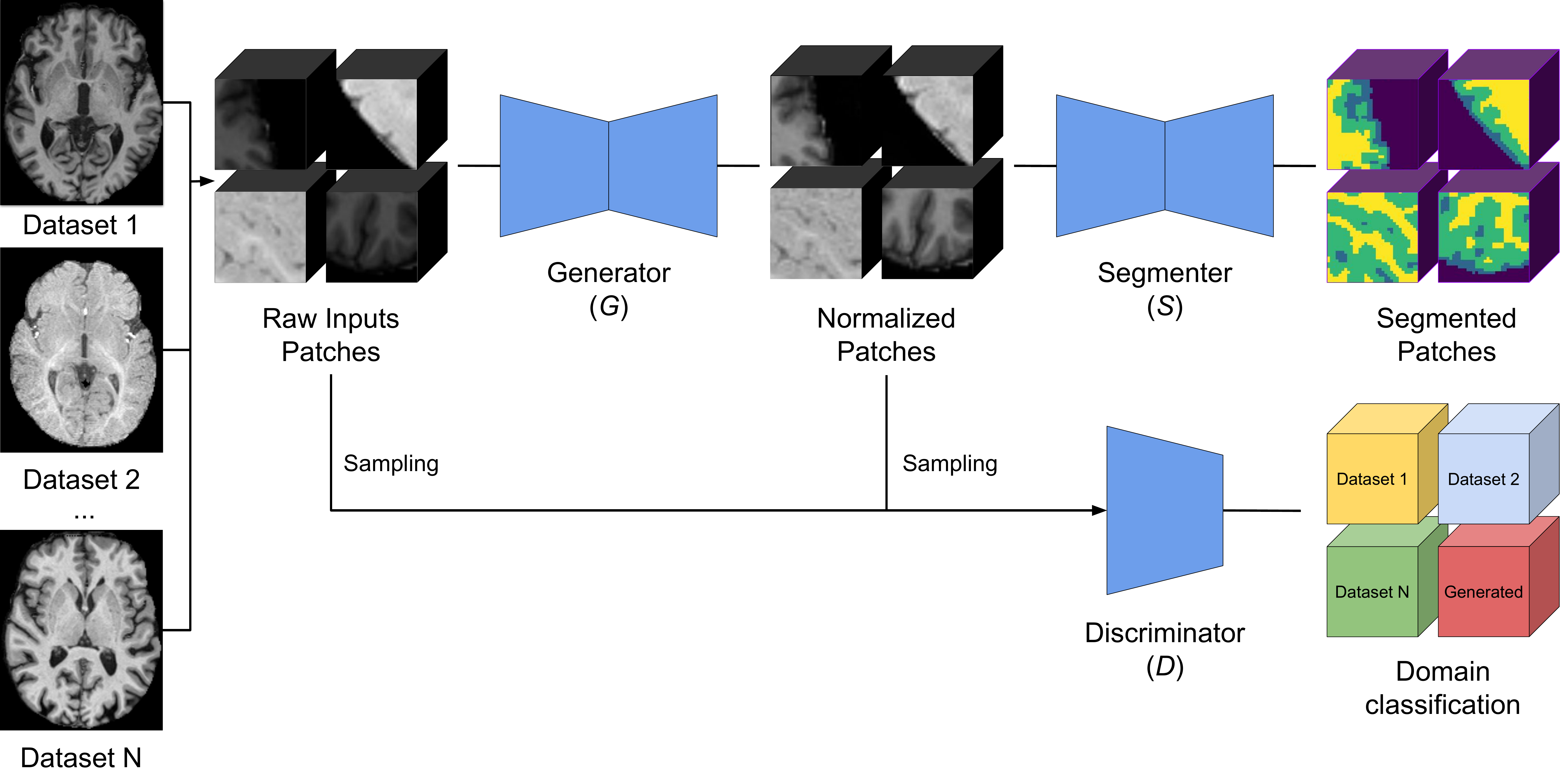}
    \caption{Proposed architecture. A first FCN generator network (G) takes a non-normalized patch and generates a normalized patch. The normalized patch is input to a second FCN segmentation network (S) for proper segmentation. Discriminator (D) network apply the constraint of realism on the normalized output. The algorithm learns the optimal normalizing function based on the observed differences between input datasets.}
\end{center}
\end{figure*}

Recently, the problems of image normalization and learned pre-processing of medical images have generated a growing interest. In \cite{RN23}, it was shown that two consecutive fully-convolutional deep neural networks (FCN) can normalize an input prior to segmentation. However, the intermediary synthetic images produced by forward pass on the first fully-convolutional network lack interpretability as there is no constrain on the realism of produced images (see Fig. \ref{fig1}). Also, a limitation of this previous work is the separate processing of 2-D slices of volumetric data, which does not take into account the valuable 3-D context of this data. The study in \cite{Onofrey2019} analyzed multi-site training of deep learning algorithms and compared various traditional normalization techniques such as histogram matching \cite{Shapira2013MultipleHM} or Gaussian standardization across different datasets. However, these standard normalization techniques are not learned for a specific task, certainly leading to suboptimal results compared to a task-driven normalization approach. Gradient reversal layer based domain adaptation and layer-wise domain-adversarial networks are used in \cite{Ciga2019} but this method is limited to two domains.

This paper extends medical image normalization so that the normalized images remain realistic and interpretable by clinicians. Our method also leverages information from multiple datasets by learning a joint normalizing transformation accounting for large image variability. We propose a task-and-data-driven adversarial normalization technique that constrains the normalized image to be realistic and optimal for the image segmentation task. Our approach exploits two fully-convolutional 3-D deep neural networks \cite{RN10}. The first acts as a normalized image generator, while the second serves as segmentation network. This paper also includes a 3-D discriminator \cite{RN12} network that constrains the generator to produce interpretable images. Standard generative adversarial networks (GANs) \cite{Goodfellow2014} aim at classifying generated images as fake or real. Our discriminator rather acts as a domain classifier by distinguishing images between all input domains (i.e. a dataset which has been sampled with a specific protocol at a specific location) including an additional ``generated'' one. Hence, produced images are both realistic and domain invariant. The parameters of all three networks are learned end-to-end. 

Our contributions can be summarized as follows: 1) an adversarially-constrained 3-D pre-processing and segmentation technique using fully-convolutional neural networks which can train on more than one dataset, 2) a learned normalization network for medical images which produces images that are realistic and interpretable by clinicians. The proposed method yields a significant improvement in segmentation performance over using a conventional segmentation model trained and tested on two different data distributions which haven't been normalized by a learning approach. To the best of our knowledge this is the first work using purely task-and-data driven medical image normalization while keeping the intermediary image medically usable.

\begin{table*}[t!]
  \centering
  \begin{small}  
  \begin{tabular}{r*{6}{c}}
    \toprule
    &  &  & & \multicolumn{3}{c}{Dice} \\
     \cmidrule(lr){5-7}
    Exp. \# & Method & Train dataset & Test dataset & CSF & GM & WM \\
    \midrule
    1 & No adaptation & iSEG & iSEG & 0.906  & 0.868 & 0.863 \\
    2 & No adaptation & MRBrainS & MRBrainS & 0.813 & 0.789 & 0.839 \\
    3 & No adaptation, Cross-testing & iSEG & MRBrainS & 0.401 & 0.354 & 0.519  \\
    4 & No adaptation, Cross-testing & MRBrainS & iSEG & 0.293 & 0.082 & 0.563  \\
      \midrule
    5 & Standardized & iSEG + MRBrainS & iSEG + MRBrainS & 0.849 & 0.808 & 0.809 \\
    & & Standardized & Standardized & & & \\ 
    6 & Without constraint & iSEG + MRBrainS & iSEG + MRBrainS & 0.834  & 0.859 & 0.885 \\
    7 & {\textbf{Adversarially normalized (ours)}} & iSEG + MRBrainS & iSEG + MRBrainS & \textbf{0.919} & \textbf{0.902} & \textbf{0.905} \\
    \bottomrule
  \end{tabular}
  \end{small}
  \caption{Dice score in function of the model architecture and data. The proposed method yielded a significant performance improvement over training and testing on single-domain or on standardized inputs.}\label{tlc}
\end{table*}

\section{Method}

Let $\xx \in \real^{|\img|}$ be a 3-D image, where $\img$ is the set of voxels, and $\yy \in \{0,1\}^{|\img| \times C}$ be its segmentation ground truth with pixel labels in $\{1, \ldots, C\}$. The training set $\data = \{(\xx_i, \yy_i, z_i) \, | \, i = 1, \ldots, M\}$ contains $M$ examples, each composed of an image $\xx_i$, a manual expert segmentation $\yy_i$ and an image domain label $z_i \in \{1, \ldots, K\}$. As shown in Fig. 2, the proposed model is composed of three networks. The first network $G$ is a fully-convolutional network. A 3-D U-Net architecture, without loss of generality, has been chosen for its simplicity. This network transforms an input image $\xx$ into a cross-domain normalized image $\xnorm = G(\xx)$. The second network $S$, which is also a 3-D FCN, receives the normalized image as input and outputs the segmentation map $S(\xnorm)$. The third network $D$ is the discriminator which receives both raw images and normalized images as input and predicts their domain. Network $D$ learns a $(K\!+\!1)$-class classification problem, with one class for each raw image domain and a $(K\!+\!1)$-th class for generated images of any domain. As mentioned before, the discriminator is used to ensure that images produced by $G$ are both realistic and domain invariant.

The three networks of our model are trained together in an adversarial manner by optimizing the following loss function:
\begin{align}\label{eq:total_loss}
& \min_{G,S} \max_{D} \ \loss(G, S, D) \ = \ \sum_{i=1}^{M} \lossSeg \big(S(G(\xx_i)), \yy_i\big)  \\[-1mm] 
& \quad - \ \lambda\left[\sum_{i=1}^{M} \left( \lossDis\big(D(\xx_i), z_i\big) \, + \, \lossDis\big(D(G(\xx_i)), \mr{fake}\big)\right )\right]\nonumber. % adding a comma
\end{align} 
For training the segmentation network, we use the weighted Dice loss defined as
\begin{equation}
\lossSeg(\sss,\yy) \ = \ 1 \, - \, \frac{\epsilon \, + \, 2 \sum_{c} w_c \sum_{v \in \img} s_{v,c} \cdot y_{v,c}}{\epsilon \, + \, \sum_{c} w_c \sum_{v \in \img} (s_{v,c} +  y_{v,c})}
\end{equation}
where $s_{v,c} \in [0, 1]$ is the softmax output of $S$ for voxel $v$ and class $c$, and $\epsilon$ is a small constant to avoid zero-division. For the discriminator classification loss, we employ we use the standard log loss. Let $p_D$ be the output class distribution of $D$ following the softmax. For raw (unnormalized) images, the loss is given by
\begin{equation}
    \lossDis\big(D(\xx), z\big) \ = \ - \log \, p_D(Z = z \, | \, \xx).
\end{equation}
On the other hand, in the case of generated (normalized) images, the loss becomes
\begin{align}
    \lossDis\big(D(G(\xx)), \mr{fake}\big) & \ = \
        - \log \, p_D(Z = \mr{fake} \, | \, G(\xx)) \\
        & \ = \ - \log \big[1 - p_D(Z \leq K \, | \, G(\xx))\big]\nonumber
\end{align}
As in standard adversarial learning methods, we train our model in two alternating steps, first updating the parameters of $G$ and $S$, and then updating the parameters of $D$.

\section{Experiments and results}
\label{sec:pagestyle}

\subsection{Data}
To evaluate the performance of our method, two databases have been carefully retained for their drastic difference in intensity profile and nature. The first one, iSEG \cite{Wang2019}, is a set of 10 T1 MRI images of infant. The ground truth is the segmentation of the three main structures of the brain: white matter (WM), gray matter (GM) and cerebrospinal fluids (CSF), all three being critical for detecting abnormalities in brain development. Images are sampled into an isotropic 1.0 mm$^3$ resolution. 
The second dataset is MRBrainS \cite{Mendrik2015} which contains 5 adult subjects with T1 modality. The dataset also contains the same classes as ground truth. Images were acquired following a voxel size of 0.958\,mm $\times$ 0.958\,mm $\times$ 3.0\,mm.

While both datasets have been acquired with a 3 Tesla scanner, iSEG is a set of images acquired on 6-8 month-old infants while MRBrainS is a set of adult images. This implies strong structural differences. Moreover, iSEG is particularly challenging because the images have been acquired during subject's isointense phase in which the white matter and gray matter voxel intensities greatly overlap, thus leading to a lower tissue contrast. This lower contrast is known to misleads common classifiers. 

\subsection{Pre-processing and implementation details}

Only T1 images were used in our experiments. Since images in MRBrainS dataset are a full head scan, skull stripping was performed using the segmentation map. Resampling to isotropic 1.0\,mm$^3$ as been done to match iSEG sampling. Overlapping patches of $32^3$ voxels centered on foreground were extracted from full volumes with stride of $8^3$ which yielded 27,647 3-D patches in total for all training images. Each training took 8 hours using distributed stochastic gradient descent on two NVIDIA RTX 2080 Ti GPUs. 

\begin{table}[ht!]
  \centering
  \begin{small}
  \begin{tabular}{r*{6}{c}}
    \toprule
    &  & Input data & Normalized images \\
    \midrule
    & JSD & 3.509 & 3.504 \\
    \bottomrule
  \end{tabular}
  \end{small}
  \caption{Jensen-Shannon divergence (JSD) of input and normalized images from the generator. A lower value corresponds to more similar distributions.}\label{tlc}  
\end{table}

\begin{figure}[ht!]
    \includegraphics[width=\linewidth]{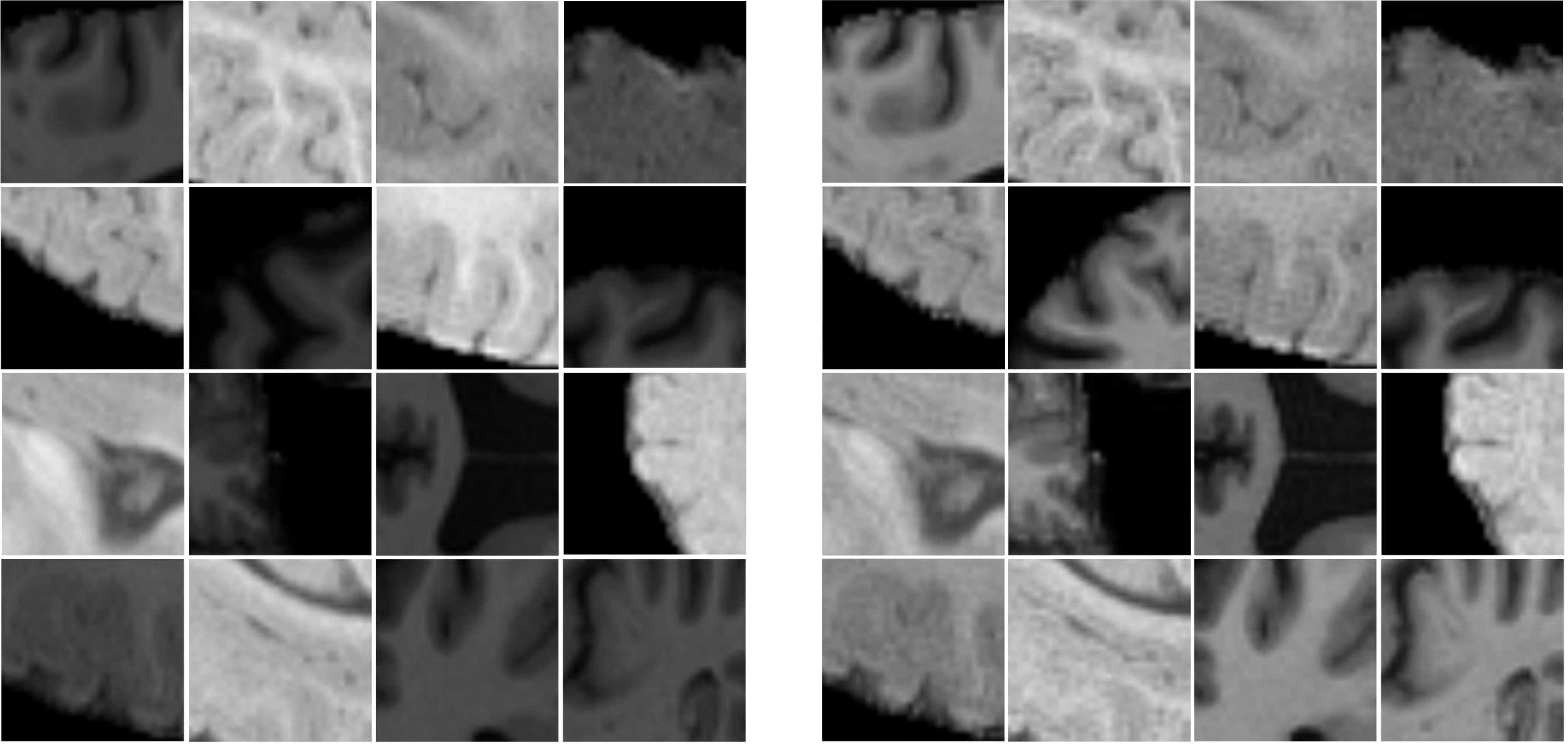}
    \caption{Mixed iSEG and MRBRainS inputs (\textbf{left}) and the generated images with adversarial normalization with $\lambda\!=\!1.0$ (\textbf{right}). Notice the the improved homogeneity of intensities in the normalized images, making their analysis easier.}\label{normalized}
\end{figure}

\begin{figure*}[ht!]
    \begin{footnotesize}
    \shortstack{\includegraphics[width=.33\linewidth]{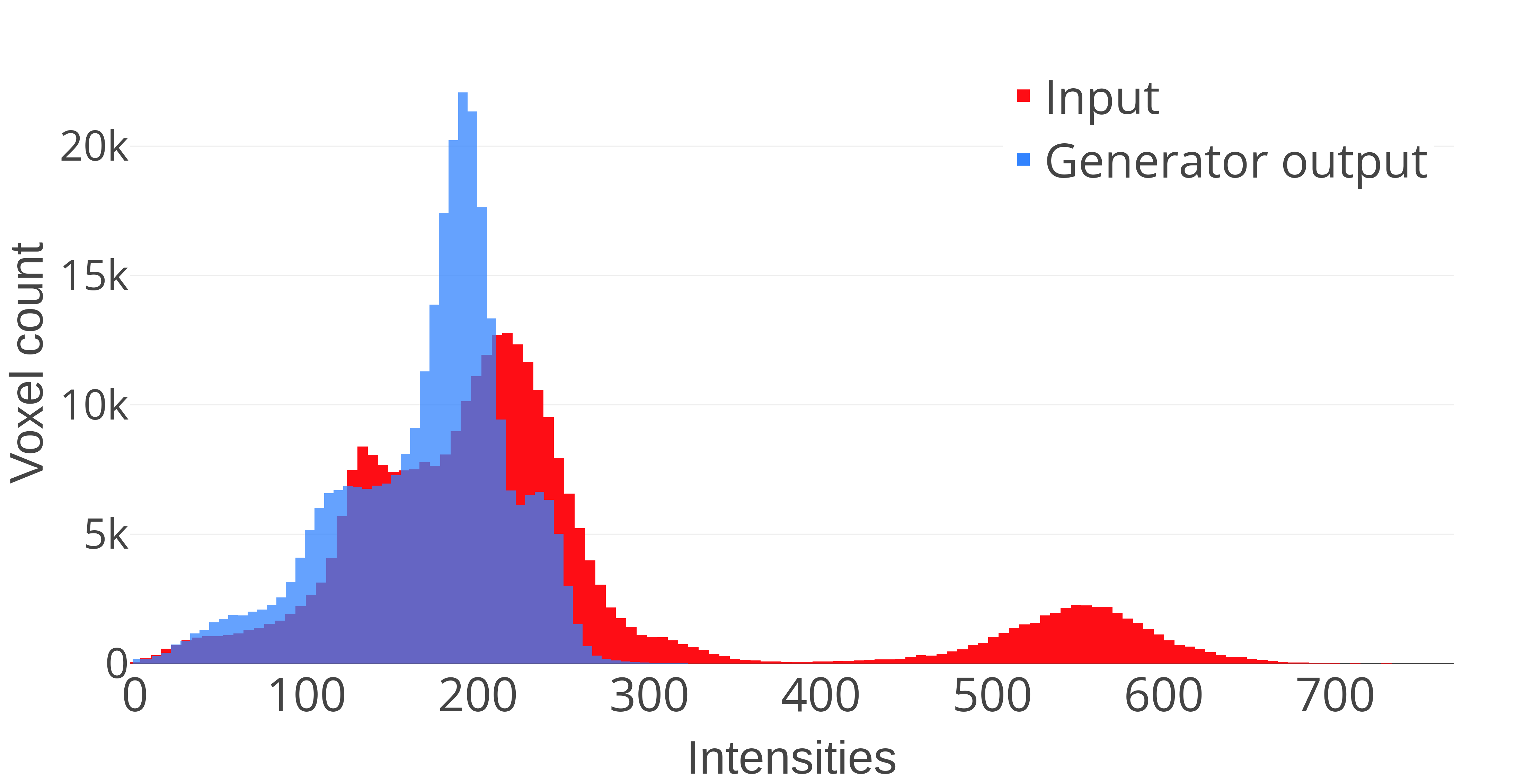} \\ GM}
    \shortstack{\includegraphics[width=.33\linewidth]{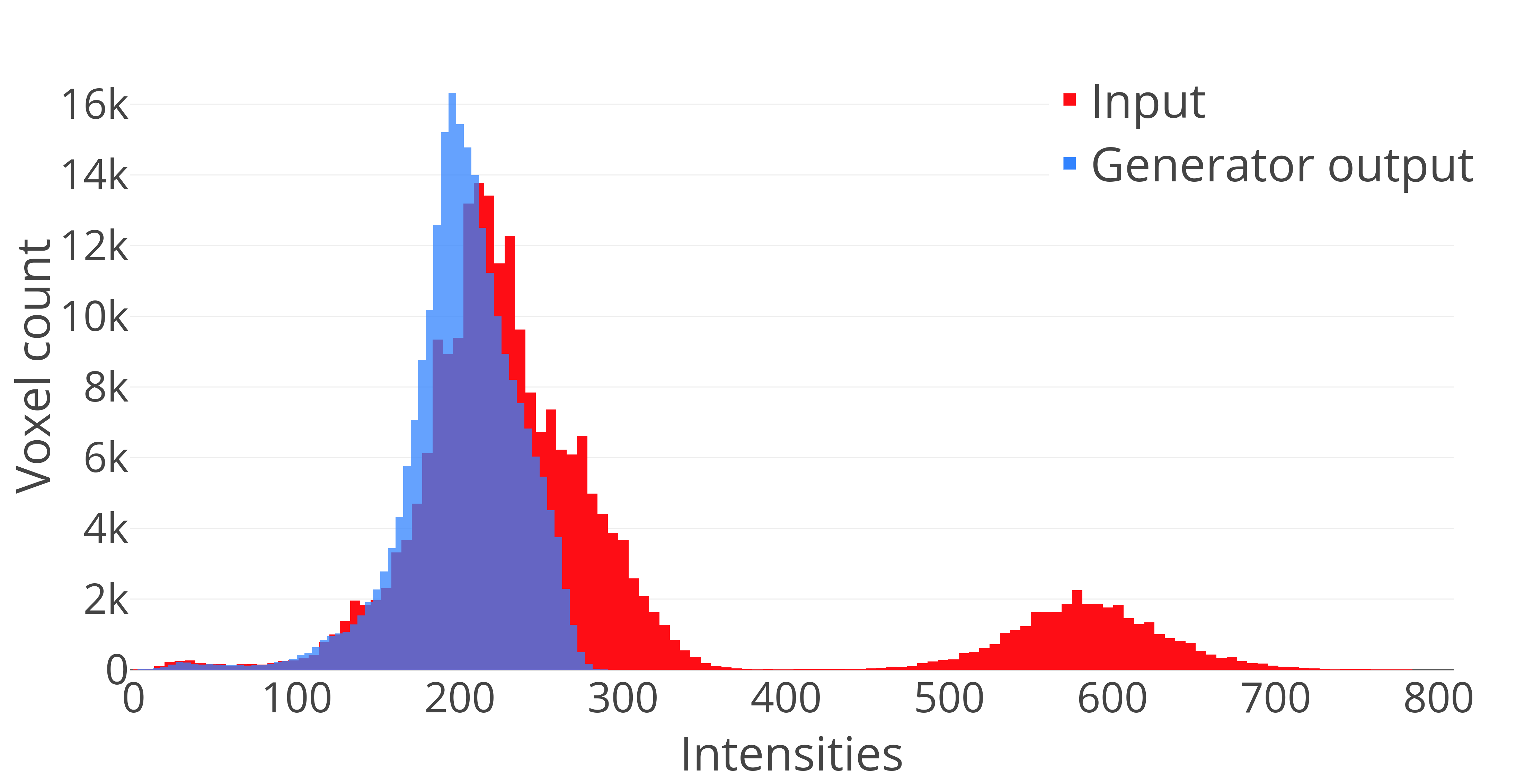} \\ WM}
    \shortstack{\includegraphics[width=.33\linewidth]{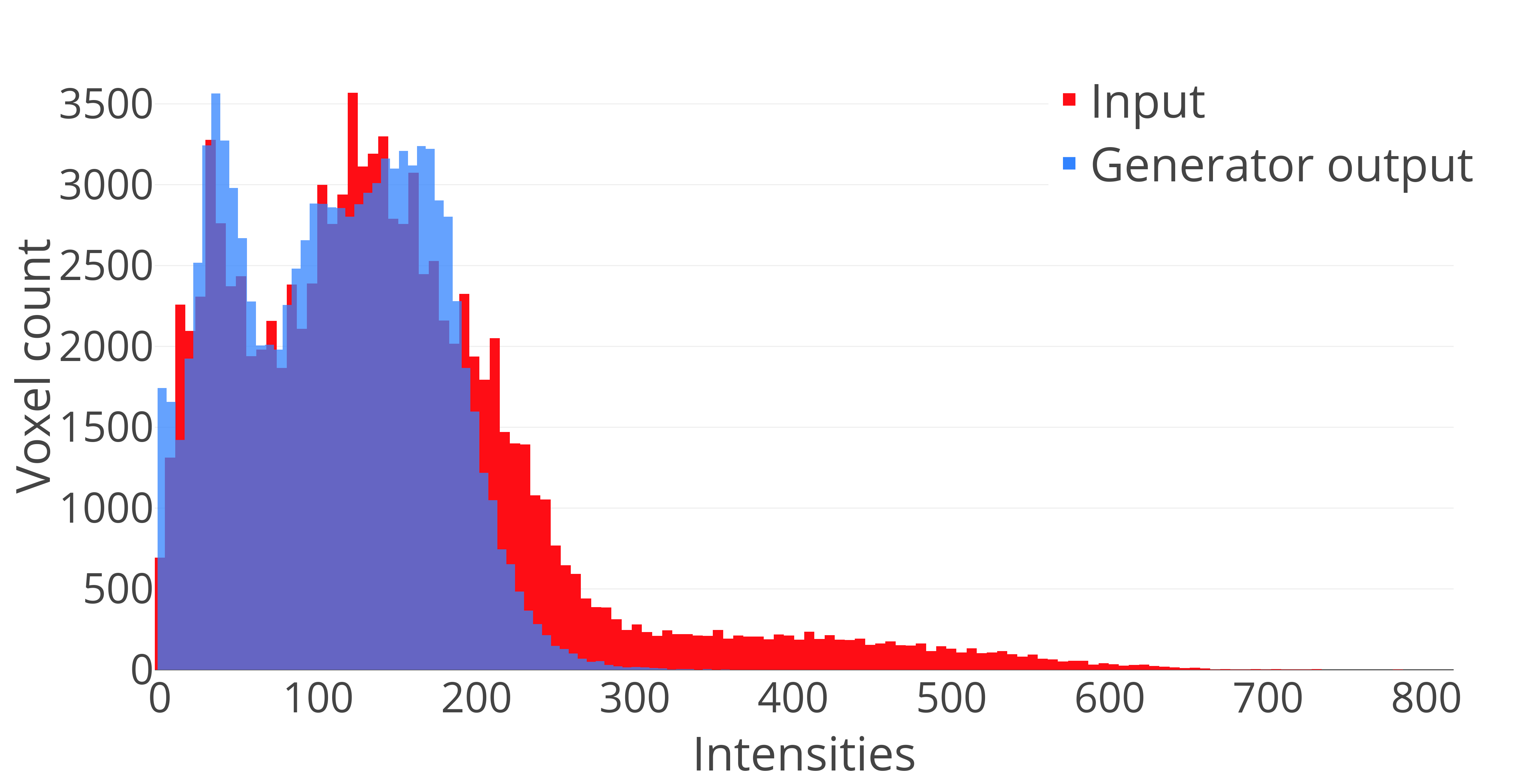} \\ CSF}    
    \end{footnotesize}      
    \caption{Histograms of generator outputs (blue) and unnormalized inputs (red). The intensity range of the generated images for grey matter (GM), white matter (WM) and CSF is more compact, showing a reduced variance in voxel intensities.}\label{histograms}
\end{figure*}
    
\subsection{Experiment setup}

We split 60\% of the patches for training, while the other 40\% are split in half for validation and testing. Split is done following a stratified shuffle split strategy based on center voxel class. Seven different experimental settings were tested. Baselines for segmentation performance on respectively training and testing on each dataset are done (Exp. 1-2 in Table 1). Cross-domain testing is then done for both dataset (Exp. 3-4). For comparison purpose, we trained a segmentation network with both datasets on which we applied Gaussian standardization (Exp. 5). We also trained two pipelined FCN networks with both datasets at the same time (Exp. 6). We then trained using our method with both datasets (Exp. 7). For each experiment involving the generator network, a Mean-Square Error (MSE) loss is used to initialize weights and produce an output close to real input. Optimum is reached after three epochs. Segmentation Dice loss and discriminator's cross-entropy loss is then added at the beginning of the fourth epoch. Each experiment has been trained with a Stochastic Gradient Descent optimizer with momentum of 0.9 and weight decay of 0.1. All networks have been initialized using Kaiming initialization \cite{He2015}. Generator uses a learning rate of $0.00001$ while segmentation and discriminator networks use $0.0001$. A learning rate scheduler with patience of 3 epochs on validation losses reduces the learning rate of all optimizers by a factor of 10. No data augmentation was applied.

\subsection{Normalization performance}

To evaluate the normalization performance, we used the Jensen-Shannon divergence (JSD) between intensity histograms of images in the validation set. JSD measures the mean KL divergence between a distribution (i.e., histogram) and the average of distributions. Table \ref{tlc} gives the JSD between input images and between images normalized by the generator. We see a decrease in JSD for normalized images showing that the intensity profiles of generated images are more similar to each other. The normalization effect of our method can be better appreciated in Fig. \ref{histograms}. This figure shows a narrower distribution that is more centered around a single mode therefore reducing the intra-class variance and increasing segmentation accuracy. Another benefit of our method is the contrast enhancement it provides to the generated images (Fig. \ref{normalized}). This is mainly brought by our task-driven approach, where minimizing the segmentation loss helps at increasing the contrast along region boundaries. 

\subsection{Segmentation performance}

Our method relies on the segmentation task while performing online normalization. Since the Dice loss is being used, structural elements are kept when generating the intermediate image, while cross-entropy aims at keeping the global features of the image while reducing the differences across domains. The main advantage of our method is the ability to train on drastically different datasets regarding to structures and intensity distributions while still maintaining a good segmentation performance. We are effectively performing segmentation of structures on both adults and infant brains at the same time while still achieving 90.8\% of mean Dice score across all classes. This demonstrates the relevance of adding adversarial normalization, increasing the Dice score of up to 59.6\% up in mean segmentation performance over all classes against training on a single dataset and testing on the other one. As seen in Fig. 3, our method is also able to normalize the images while maximizing the segmentation and keeping the image interpretable and realistic. This is achieved by the discriminator's loss which aims at minimizing the cross-entropy between real inputs and generated input up to the point it cannot differentiate anymore from which domain (real iSEG input, real MRBrainS input or generated input) the generator's output comes from.

\section{Conclusion}

We proposed a novel task-and-data-driven normalization technique to improve a segmentation task using two datasets. Our method drastically improves performance on test sets by finding the optimal intermediate representation using the structural elements of the image brought by Dice loss and global features brought by cross-entropy. We believe this work is an important contribution to biomedical imaging as it would unlock the training of deep learning models with data from multiple sites, thus reducing the strain on data accessibility. This increase in data availability will help computing models with higher generalization performance. We also demonstrated that it is still possible with our method to train with different datasets while one of these is particularly difficult because of the lower image contrasts. Future work will aim at using a better model to act as a discriminator or using other training methods such as Wasserstein GAN \cite{Arjovsky2017} or CycleGAN \cite{Zhu2017} to compare performance difference on the image normalization. Our method is also theoretically capable of handling more than two datasets, finding automatically a common intermediate representation between all input domains and adapting voxel intensities to maximize the segmentation performance. Further work will also demonstrate the architecture's performance on new datasets and on the gain this task-driven normalization can provide in case of segmentation tasks with greatly imbalanced data such as brain lesion segmentation.

\medskip
\noindent
\textbf{Acknowledgment} -- This work was supported financially by the Research Council of Canada (NSERC), the  Fonds  de Recherche du Quebec (FQRNT), ETS Montreal, and NVIDIA with the donation of a GPU.

\vfill
\pagebreak

\bibliographystyle{IEEEbib}
\bibliography{refs}

\end{document}